# Determination of the most representative descriptor among a set of feature vectors for the same object

Dmitry Pozdnyakov
*Oxagile, Dzerginskogo av.57, Minsk 220089, Belarus*
*E-mail: pozdnyakov@tut.by; dmitry.pozdnyakov@oxagile.com*

**Abstract:** On an example of solution of the face recognition problem the approach for estimation of the most representative descriptor among a set of feature vectors for the same face is considered in present study. The estimation is based on robust calculation of the mode-median mixture vector for the set as the descriptor by means of Welsch/Leclerc loss function application in case of very sparse filling of the feature space with feature vectors.

**Introduction**

The interest to the deep learning methods for solution of such problems of the field of computer vision like object detection and recognition on images and video has increased dramatically for the last several years. One of the problems is detection of people faces on images and video [1–3] as well as their digital (post-)processing and recognition [4–15]. Person recognition is distinction of different people by means of comparison of their key face features. If not to take into account papers in which the comparison procedure is not formalized, and it is also product of neural network training, see as example [8], then the classical approach of the mentioned problem solution is a face similarity estimation by cosine similarity for the feature vectors corresponding to the faces [8–10]. The vectors are elements of *N*-dimensional feature space $\mathbb{R}^N$. The space itself is filled with the vectors in the process of neural network training, and their spatial distribution depends on the neural network architecture and loss function type [11–15].

One of the tasks of a general recognition problem is a limited discriminativity of the feature vectors for objects of the same class. The method of its solution is well known [11–15]: to find such a neural network architecture, but mainly loss functions, for which from the one hand different images of the same object form the most compact cluster of the vectors in the feature space, and from the other hand a distance between different clusters corresponding to different objects is large as possible for all the objects of the same class. In due time the authors of paper [13] achieved the greatest success in the problem solution by consideration of the hypersphere $S^{N-1}$ instead of the hyperspace $\mathbb{R}^N$. Further general development of their approach was presented in papers [14, 15]. In the framework of the approach the length of a feature vector extracted by the neural network directly corresponds to the contrast and sharpness of the face image being processed, that can be elementary proved by analysis of expressions describing the neural network architecture on the mathematical level. At the same time the position of descriptor, that is a feature vector normalized to the unity length, in the feature space determines the unique features balance for every unique face along with the features deviation due to various emotions on the face, different tilts and/or rotations of the head as well as different garment and decoration elements in the neural network receptive field (field of view). The global similarity of the people faces is accordingly determined by the descriptors cosine similarity which numerically coincides with the value of their dot product [8–15].

Another task of the recognition problem is reliable search for the most similar face image by descriptor to the compared face image among all the faces images from the corresponding dataset. And if all the images for all the persons in the dataset are high quality (frontal photos with high-resolution and without occlusions) then comparison process passes with a high reliability because multidimensional Gaussian(-like) distributions of the descriptors on the hypersphere look like the compact clusters with a very low degree of mixture [13–15]. But in case of presence of statistical outliers in the descriptor distributions, corresponding to the low quality images (low-resolution photos, photos with occluded or profile faces, and even other people photos), the search reliability drops out significantly. The fact is that the outliers dislocate the estimated positions of clusters centers, which are calculated through the mathematical expectation, relative to the most likelihood positions of the clusters centers — the modes of Gaussian(-like) distributions of the descriptors.

In due time the method of solution of the problems like that has been proposed in the paper [16]. Instead of classical calculation of the mathematical expectation equivalently for all the elements of the set with weighting factors equal to each other the elements are considered in this method like non-equivalent and the weighting factors themselves are self-consistently estimated in the process of iterative calculation of the mathematical expectation and dispersion.

Thus the purpose of the manuscript is development of algorithm for estimation of the central descriptor position for corresponding set of feature vectors which is robust to the statistical outliers and allows the most representative descriptor of the set to be found additionally. The developments of paper [16] can be used with necessary adaptations and generalizations as a mathematical basis of the algorithm.

**Theory**

Let's consider a set of descriptors $\mathbf{v}_k = (v_1, ..., v_N)_k^T$ ($k = 1, ..., K; K \geq 2$) characterizing the same object which are the feature vectors with unity lengths ($|\mathbf{v}_k| = 1$) from the $N$-dimensional feature space $\mathbb{R}^N$. The center of cluster created by them on the ($N - 1$)-dimensional hypersphere $S^{N-1}$ [13] is determined quite trivially through equality

$$\mathbf{v}_{cl} = \arg\min_{\mathbf{w}} \left( \sum_k \arccos^2 (\mathbf{v}_k \mathbf{w}) \right) \quad (1)$$

under the condition that $\mathbf{w}$ belongs to the hypersphere $S^{N-1}$ ($|\mathbf{w}| = 1$). In fact, Eq. (1) defines the cluster central vector $\mathbf{v}_{cl}$ by dispersion minimization for squared angular distances between it and the cluster vectors on the hypersphere. By other words $\mathbf{v}_{cl}$ is the average descriptor for the considered cluster of feature vectors from the sub-space $S^{N-1}$ of the space $\mathbb{R}^N$.

The most representative cluster descriptor $\mathbf{v}_{mr}$ is a vector which is closest to the cluster central vector $\mathbf{v}_{cl}$ on the considered ($N - 1$)-dimensional hypersphere. That is

$$\mathbf{v}_{mr} = \arg\max_k (\mathbf{v}_{cl} \mathbf{v}_k). \quad (2)$$

It should be noted that the direct application of Eq. (1) to find $\mathbf{v}_{cl}$ is associated with significant computational difficulties. Whereas it is not really necessary to use Eq. (1) to find $\mathbf{v}_{mr}$ because basing on Eqs. (1) and (2) the equality

$$\mathbf{v}_{mr} = \arg\min_k \left( \sum_{k'} \arccos^2 (\mathbf{v}_{k'} \mathbf{v}_k) \right) \quad (3)$$

is true. Moreover, median estimations for the vectors $\mathbf{v}_{cl}$ and $\mathbf{v}_{mr}$, that are more robust to outliers [16], can be obtained for the cluster elements when deviations of angular distances between $\mathbf{v}_{cl}$ ($\mathbf{v}_{mr}$) and the cluster vectors are being minimized on the hypersphere under the condition that $\arccos \in [0, \pi]$. Namely, the formulae

$$\mathbf{v}_{cl} = \arg\min_{\mathbf{w}} \left( \sum_k \arccos(\mathbf{v}_k \mathbf{w}) \right), \quad (4)$$

$$\mathbf{v}_{mr} = \arg\min_k \left( \sum_{k'} \arccos(\mathbf{v}_{k'} \mathbf{v}_k) \right) \quad (5)$$

take place.

Taking into account the fact that a non-strict equality

$$\frac{\sin(\langle\alpha\rangle)}{\cos(\langle\alpha\rangle)} \approx \frac{\langle\sin\alpha\rangle}{\langle\cos\alpha\rangle} \quad (6)$$

is true with a very high degree of accuracy at $\min_{k,k'}(\mathbf{v}_k\mathbf{v}_{k'}) > 0$ (it is very rarely $\min_{k,k'}(\mathbf{v}_k\mathbf{v}_{k'}) < 0.5$ for the real practical cases [13–15]), the vector $\mathbf{v}_{cl}$ can be rather accurately calculated with application of expressions

$$\mathbf{v}_{cl} \approx \frac{\mathbf{v}_{av}}{|\mathbf{v}_{av}|}, \quad (7)$$

$$\mathbf{v}_{av} = \arg\min_{\mathbf{w}}\left(\sum_k (\mathbf{v}_k - \mathbf{w})^2\right) \quad (8)$$

that are based not on angular dispersion estimation on the hypersphere $S^{N-1}$ but on spatial dispersion estimation in the plane hyperspace $\mathbb{R}^N$ in which the squared distance is just $L_2$-norm.

Eq. (8) can be solved analytically in the form of equalities

$$\mathbf{v}_{av} = \sum_k \omega_k \mathbf{v}_k, \quad (9)$$

$$\omega_k = 1/K. \quad (10)$$

The cluster central vector found through mathematical expectation is a very good posteriori estimation for the symmetric unimodal Gaussian(-like) distribution function top corresponding to the mode of multidimensional spatial Gaussian(-like) distribution of object features in the feature space [13–15] only in case of absence of outliers in the distribution [16]. Otherwise, either in the presence of statistical outliers or in the presence of asymmetries in the spatial distribution of feature vectors, it is necessary to apply another estimation to find the mode corresponding to the most likelihood position for the center of cluster. A better variant is estimation through the median (Eqs. (4) and (5)). But the best solution is a posteriori direct estimation of the mode. As is known [16], the problem can be solved by means of introducing of the limiting Welsch/Leclerc loss function [16–18]

$$f(x,c) = 1 - \exp\left(-\frac{x^2}{2c^2}\right) \quad (11)$$

into calculations of mathematical expectation.

As example, the system of equalities

$$\begin{cases} \omega_k^{(1)} = \dfrac{1}{K}, \\ m_q = \sum_k \omega_k^{(q)} x_k, \\ \sigma_q^2 = \sum_k \omega_k^{(q)} (x_k - m_q)^2, \\ \omega_k^{(q+1)} = \dfrac{\exp\left(-(x_k - m_q)^2/(2\sigma_q^2)\right)}{\sum_k \exp\left(-(x_k - m_q)^2/(2\sigma_q^2)\right)} \end{cases} \quad (12)$$

can be written for a scalar (one-dimensional) random value $X:\{x_k\}$ characterized by a distribution density function $\rho(x)$. In Eqs. (12) all the parameters are calculated iteratively, $q = 1, ..., Q$ ($Q \geq 2$) is an iteration number.

Let's consider the random value $X \in [0, \infty)$ under the condition that $\rho(x) = x\exp(-x)$. For such a case a priori data are the following: $\langle X \rangle = 2$, $\text{median}(X) \approx 1.678$, $\text{mode}(X) = 1$. A posteriori prediction for $m_q$ obtained by application of the iterative scheme (12) at $K = 10^9$ is presented in the figure below.

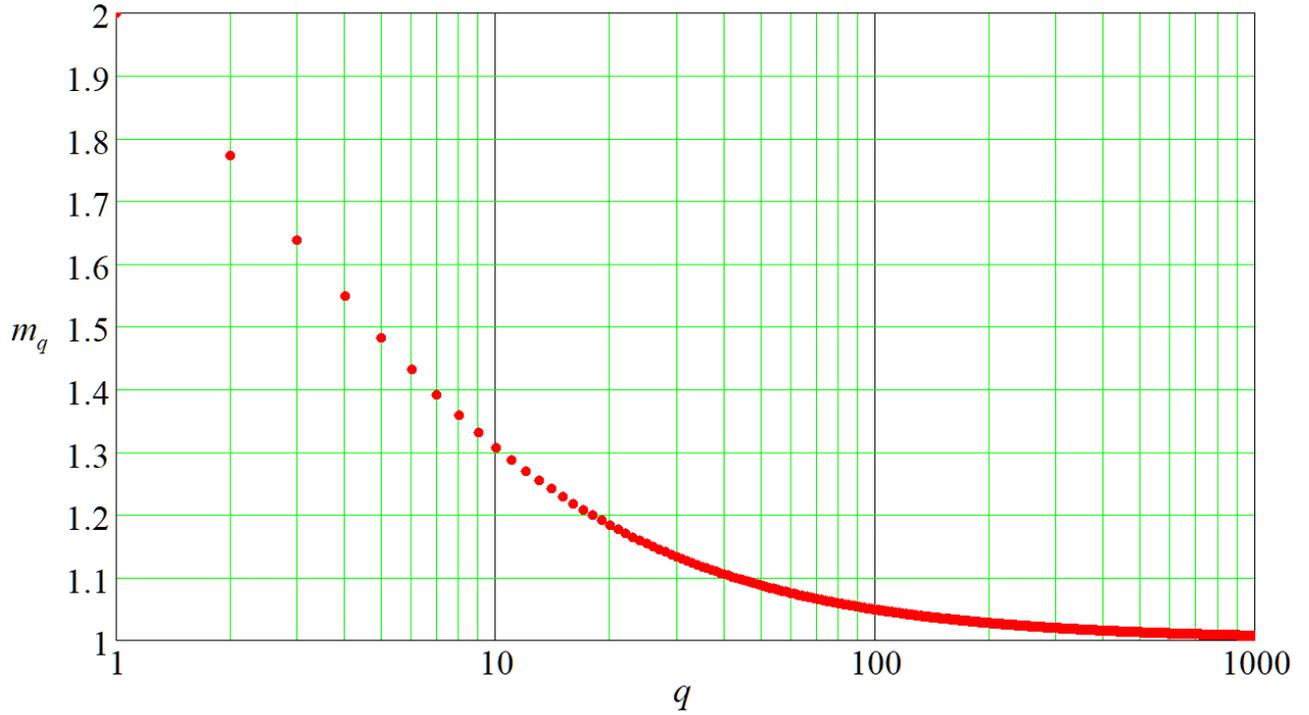

**Fig.1** – Average weighted value for $X$ depending on iteration number $q$

It follows from Fig.1 that convergence of $m_q$ from the mathematical expectation to the mode is monotonous and very slow after the first 10 iterations, and
$$m_3 \approx \mathrm{median}(X). \quad (13)$$
Moreover, the value of $Q$ is not arbitrary. The statistical reliability of the estimation of $m_q$ leads to the dependence of $Q$ on $K$. From the one hand a "complete" convergence of $m_q$ to the mode takes place for large values of $K$ at $Q = \lfloor \sqrt{K} \rfloor$ with squared logarithmic decrease of estimation reliability for $m_q$ in comparison with $m_1$. That is
$$\sigma_{\lfloor \sqrt{K} \rfloor} \sim C \ln^2(K) \sigma_1. \quad (14)$$
From the other hand at $Q = \lfloor \log_2(K+1) \rfloor$ there is a good accuracy in estimation of $m_Q$ in the form of
$$\sigma_{\lfloor \log_2(K+1) \rfloor} \sim C \ln^{1/2}(K) \sigma_1, \quad (15)$$
but with untimely iterations stopping. Therefore, to calculate the optimal number of iterations it is recommended to apply the formula
$$Q = \lfloor K^{1/4} \log_2^{1/2}(K+1) \rfloor. \quad (16)$$
In any case a posteriori reliability of the median estimation for $K \geq 9$ is not worse than the estimation reliability for the mathematical expectation. In particular, a non-strict equality
$$\sigma_3 \approx \sigma_1 \quad (17)$$
is true with a high degree of accuracy.

Due to the analysis of a number of various unimodal distributions $\rho(x)$ it has been found out that all the conclusions above (Fig.1, Eqs. (13) – (17)) are very general.

Direct generalization of the iterative scheme (12) to multidimensional case is practically impossible because all the real sets of feature vectors is not representative from the statistical point of view. For example, the $N$-dimensional space must contain at least $9^N$ feature vectors even for direct multidimensional estimation of the median, but it is not impossible by virtue of $N \geq 128$. However, the problem can be circumvented by using projective approximation. In such a case every coordinate is independently considered for the descriptors $\mathbf{v}_k$, and as a result the following expressions can be obtained:

$$\mathbf{v}_{cl} \approx \frac{\mathbf{v}_{md}}{|\mathbf{v}_{md}|}, \quad (18)$$

$$\mathbf{v}_{md} = (v_1, v_2, \ldots, v_i, \ldots, v_{N-1}, v_N)^T_{md} = (m_{1,Q}, m_{2,Q}, \ldots, m_{i,Q}, \ldots, m_{N-1,Q}, m_{N,Q})^T = \mathbf{m}_Q, \quad (19)$$

$$\begin{cases} \omega_{i,k}^{(1)} = \dfrac{\varepsilon_k}{\sum_k \varepsilon_k}, \\ m_{i,q} = \sum_k \omega_{i,k}^{(q)} v_{i,k}, \\ \sigma_{i,q}^2 = \sum_k \omega_{i,k}^{(q)} (v_{i,k} - m_{i,q})^2, \\ \omega_{i,k}^{(q+1)} = \dfrac{\varepsilon_k \exp\left(-(v_{i,k} - m_{i,q})^2 / (2\sigma_{i,q}^2)\right)}{\sum_k \varepsilon_k \exp\left(-(v_{i,k} - m_{i,q})^2 / (2\sigma_{i,q}^2)\right)}. \end{cases} \quad (20)$$

In Eqs. (20) $\varepsilon_k$ is an additional weighting factor characterizing the statistical importance of the descriptor $\mathbf{v}_k$ in the cluster which is equal to 1 by default if there is not any extra data. In particular, it can be determined through the person head deviation angle $\theta$ relative to the axis of camera lens symmetry (left-right turn and/or up-down tilt) to exclude the influence of profile and/or overtilted faces (potential outliers among the feature vectors) on the position of the cluster central vector. Namely, the equalities

$$\begin{cases} \varepsilon_k = \cos^p(\theta_k), & \theta_k < \pi/2, \\ \varepsilon_k = 0, & \theta_k \geq \pi/2, \end{cases} \quad (21)$$

could be applied, where $p = 2, 3, 4, \ldots$.

**Results**

Further in the tables are examples of the most representative face images from LFW dataset corresponding to the most representative descriptors. In top rows of the tables are images found by means of the first scheme with application of Eqs. (2), (16), (18) – (20) at $\varepsilon_k = 1$. In bottom rows of the tables are images found by means of the second scheme with application of Eq. (3).

As expected in most cases the first scheme allows the most representative descriptors to be better determined. The faces which express less emotions and contain less occlusions correspond to them. The face features are more clear and recognizable. For a statistically significant number of estimates both schemes give either an equivalent but another selection outcome or absolutely the same result. In some rare cases the first scheme is slightly inferior to the second one in quality of the most representative descriptor selection. It is due to a fact that in a number of specific cases the used approximations lead to rather rough estimates. In particular, both the hypersphere curvature is neglected, when instead of hypersphere segment the corresponding hyperplane $\mathbb{R}^{N-1}$ is considered in the hyperspace $\mathbb{R}^N$ (the hypersphere is as if projected on the hyperplane: Eq. (1) → Eq. (7)), and the additional projection approximation for feature vectors is applied because the vectors are distributed on the hypersphere too sparsely.

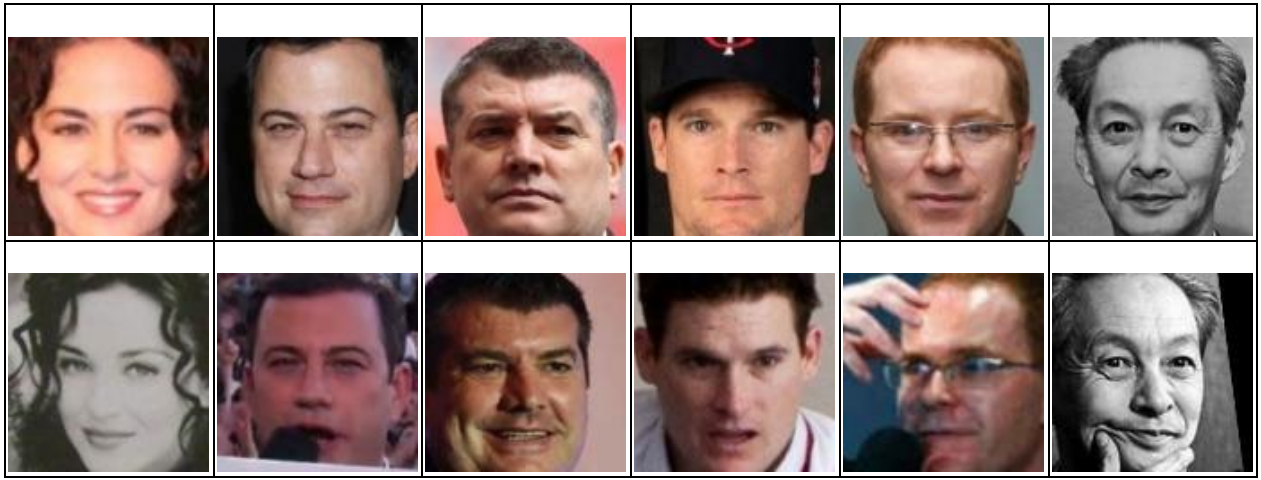

**Table 1a** – Recognition of persons in the top row is higher than in the bottom row

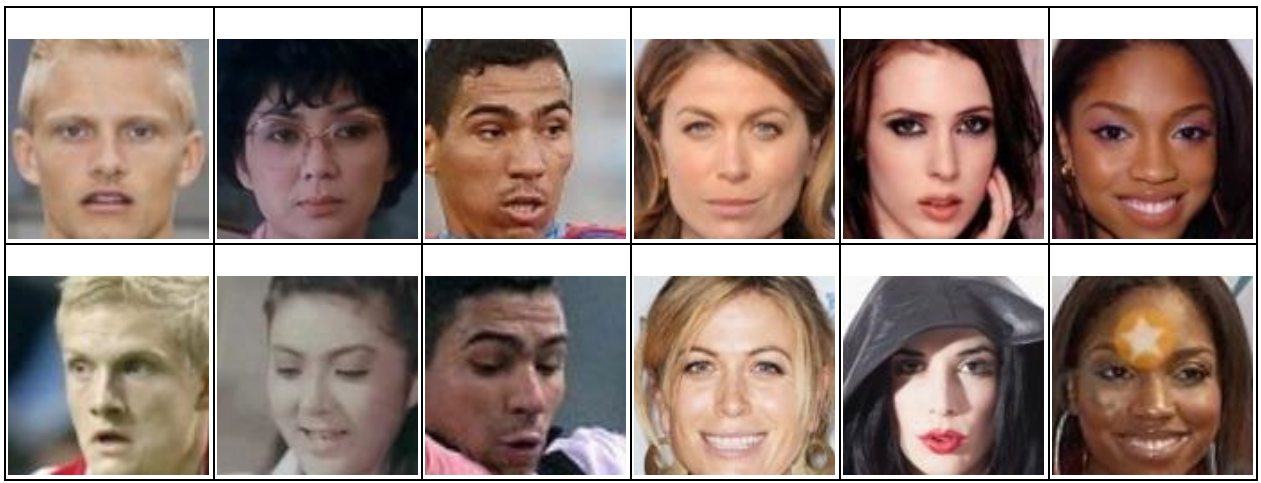

**Table 1b** – Recognition of persons in the top row is higher than in the bottom row

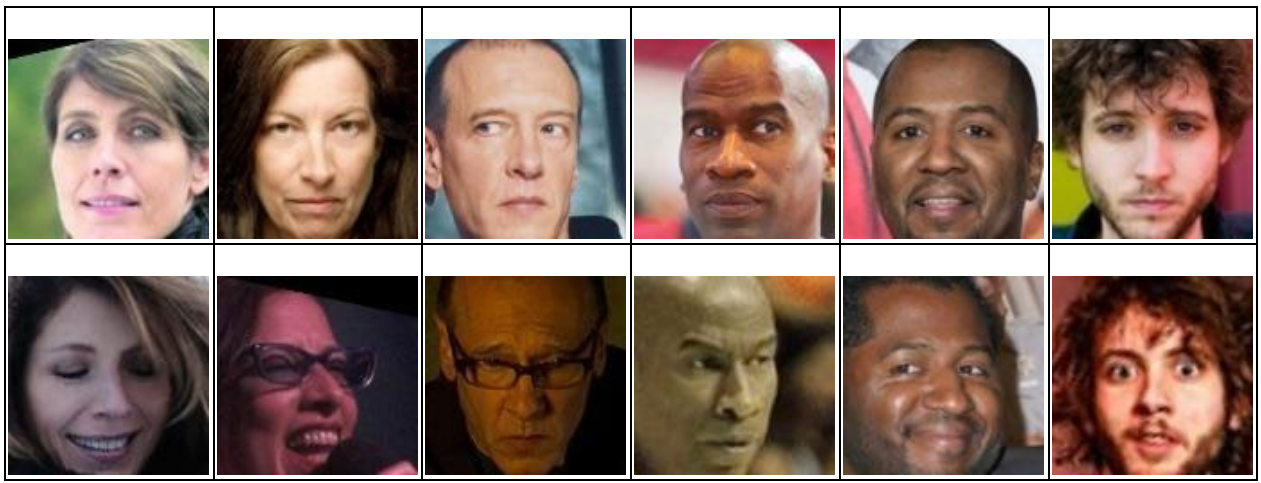

**Table 1c** – Recognition of persons in the top row is higher than in the bottom row

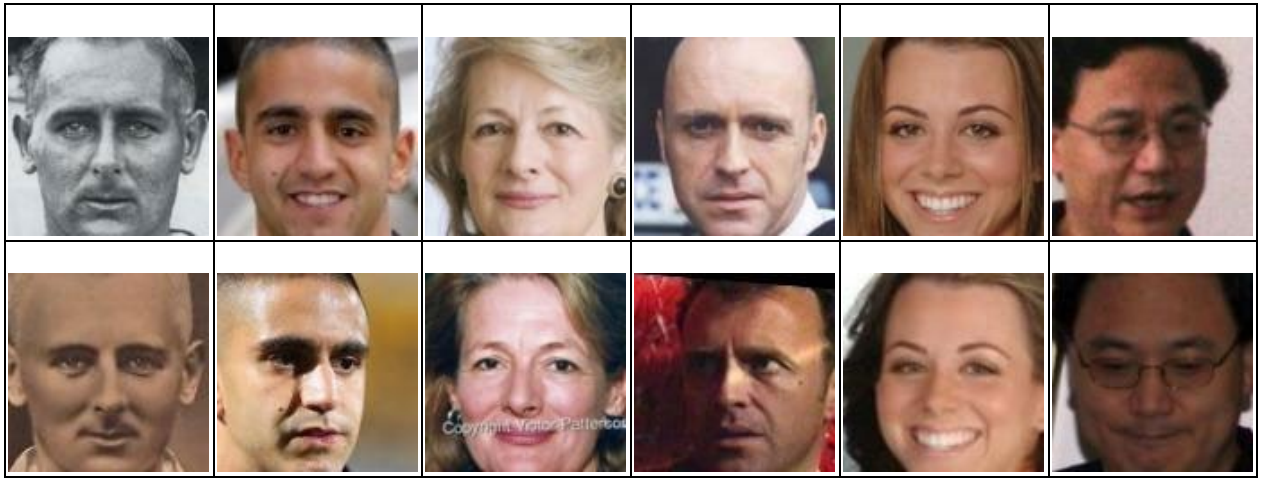

**Table 1d** – Recognition of persons in the top row is higher than in the bottom row

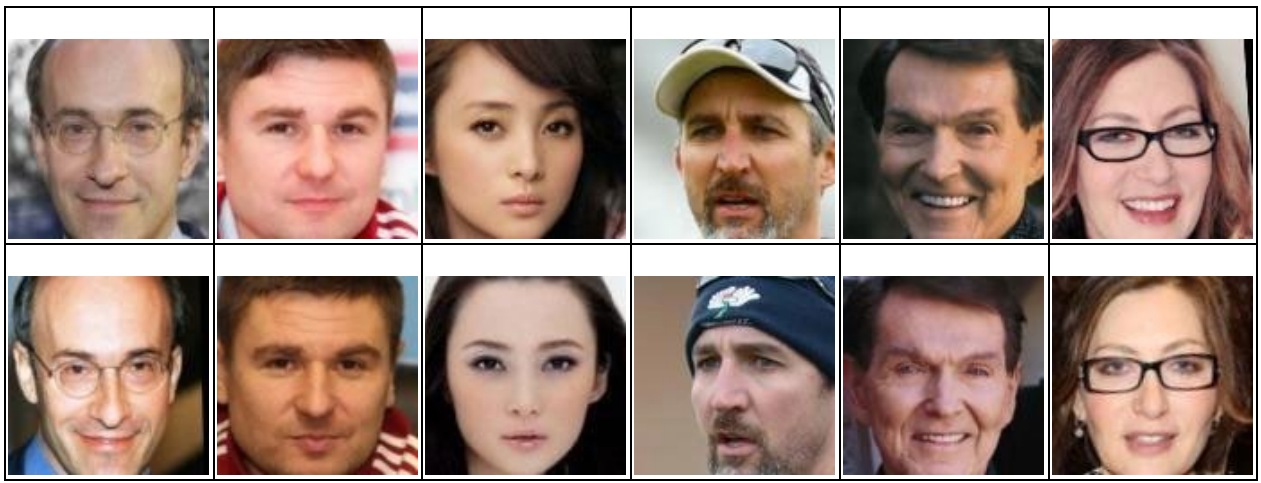

**Table 2a** – Recognition of persons is equivalent in both rows

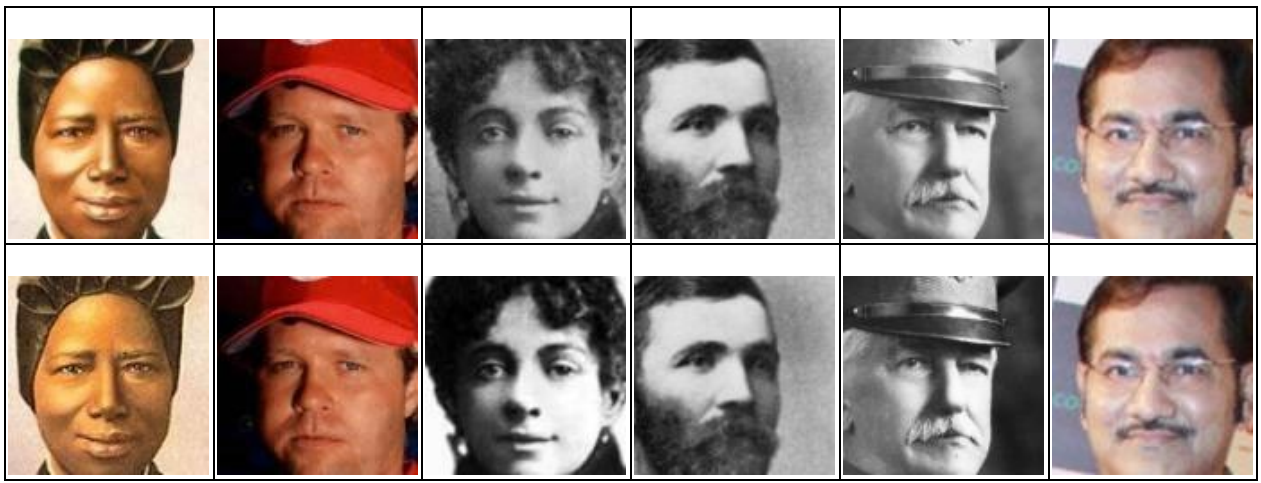

**Table 2b** – Recognition of persons is identical in both rows

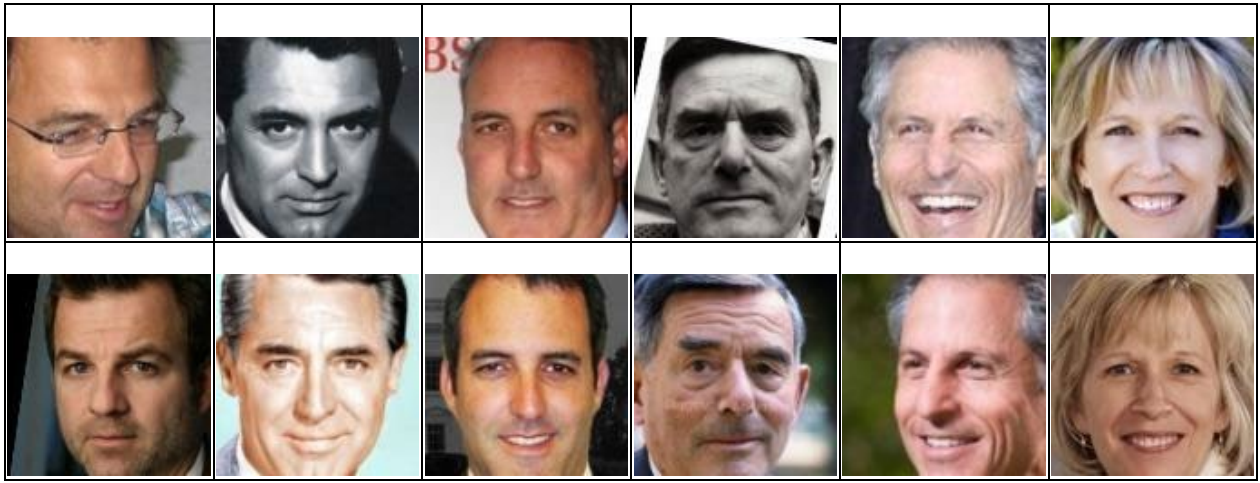

**Table 3** – Recognition of persons in the top row is lower than in the bottom row

**Conclusions**

Thus the algorithm, that is robust to statistical outliers in case of central descriptor position estimation in the feature space for corresponding descriptor set, is developed. It is shown through examples that for statistical majority of cases in spite of a number of applied approximations the considered method allows the proper position of descriptors cluster center to be predicted either better or at least not worse in comparison with the classical method of average vector calculation. Such a mode-median vector is a more accurate descriptor to characterize a whole set of face images for the same person.